\documentclass[journal, doublecolumn, 10pt]{IEEEtran}
\usepackage{setspace}
\usepackage{amsmath,amssymb,bm,amsfonts,enumerate,url}\usepackage{footnote}
\usepackage{color}
\ifCLASSINFOpdf \usepackage[pdftex]{graphicx}
\else \usepackage[dvips]{graphicx} \fi
\usepackage[linesnumbered, ruled]{algorithm2e}\SetKwRepeat{Do}{do}{while}
\usepackage[noend]{algpseudocode}
\let\oldnl\nl
\newcommand{\nonl}{\renewcommand{\nl}{\let\nl\oldnl}}
\makeatletter
\renewcommand*{\@opargbegintheorem}[3]{\trivlist
      \item[\hskip \labelsep{\bfseries #1\ #2}] \textbf{(#3)}\ \itshape}
\makeatother
\usepackage{multirow}\usepackage{cite}
\usepackage{subfig}
\usepackage{amsthm}
\def\BibTeX{{\rm B\kern-.05em{\sc i\kern-.025em b}\kern-.08em T\kern-.1667em\lower.7ex\hbox{E}\kern-.125emX}}

\usepackage{arydshln}
\usepackage{comment}

\begin{document}
\title{Zero-Touch Network on Industrial IoT: An End-to-End Machine Learning Approach}
\author{Shih-Chun~Lin, Chia-Hung Lin, and Wei-Chi~Chen
\IEEEcompsocitemizethanks{
\IEEEcompsocthanksitem Shih-Chun Lin, Chia-Hung Lin, and Wei-Chi Chen are with North Carolina State University.
}
}
\markboth{Submitted for publication in the IEEE Network}%
\maketitle

\IEEEcompsoctitleabstractindextext{%
\begin{abstract}
Industry 4.0-enabled smart factory is expected to realize the next revolution for manufacturers. Although artificial intelligence (AI) technologies have improved productivity, current use cases belong to small-scale and single-task operations. To unbound the potential of smart factory, this paper develops zero-touch network systems for intelligent manufacturing and facilitates distributed AI applications in both training and inferring stages in a large-scale manner. The open radio access network (O-RAN) architecture is first introduced for the zero-touch platform to enable globally controlling communications and computation infrastructure capability in the field. The designed serverless framework allows intelligent and efficient learning assignments and resource allocations. Hence, requested learning tasks can be assigned to appropriate robots, and the underlying infrastructure can be used to support the learning tasks without expert knowledge. Moreover, due to the proposed network system’s flexibility, powerful AI-enabled networking algorithms can be utilized to ensure service-level agreements and superior performances for factory workloads. Finally, three open research directions of backward compatibility, end-to-end enhancements, and cybersecurity are discussed for zero-touch smart factory. 





\end{abstract}
}
\maketitle
\IEEEdisplaynotcompsoctitleabstractindextext
\IEEEpeerreviewmaketitle

\section{Introduction}\label{sec_intro}



In recent years, with the development of promising technologies, such as artificial intelligence (AI), increasingly more attention from both academia and industry sides is focusing on Industry 4.0 to realize networked multi-robot systems for improved productivity \cite{Chen.2021}.
With the Industry 4.0-enabled smart factory, numerous robots, sensors, industrial robotic arms, and other industrial internet of things (IIoT) devices can be connected together to form a completed platform, providing diverse services including intelligent production line adaptation, big data analytics, manufacturing environment monitoring to further improve current productivity.
In light of this direction, various companies start to investigate or develop relative innovations. That is, the market value of smart factory will reach up to 422 billion in 2026. Specifically, Hitachi introduced their multiple AI coordination development for smart factory in May 2018, aiming at increasing the picking process efficiency in factory scenarios by integrating the control of picking robots and automated guided vehicles (AGVs) and resulting in 38\% reduced operation time compared to the original picking process. 
Another example is the human robots collaboration project announced by Mitsubishi in June 2020.
Inverse reinforcement learning algorithms \cite{IRL} for AGVs are developed so that AGVs can imitate the behavior of human workers to work with human workers without collisions and staleness for better productivity.
Via advanced AI technologies, although those works can already be used to improve the efficiency in factory scenario, one can notice that the above applications are still limited to small-scale and single-task operations.

\begin{figure}
    \centering
    \includegraphics[width = 3.4in]{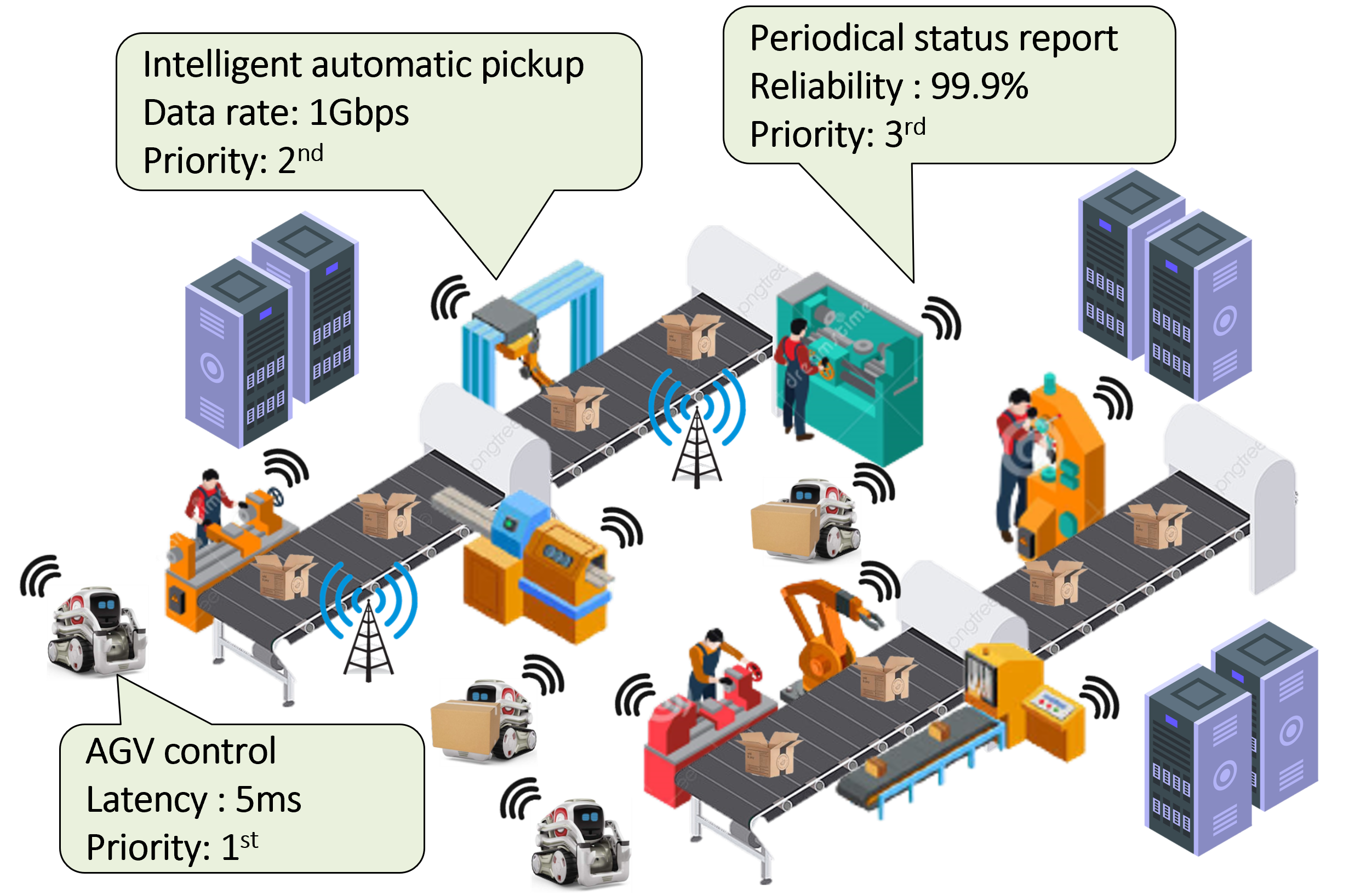}
    \caption{The smart factory scenarios.}
    \label{fig:sf_usecase}
\end{figure}

\begin{figure*}
    \centering
    \includegraphics[width=\textwidth]{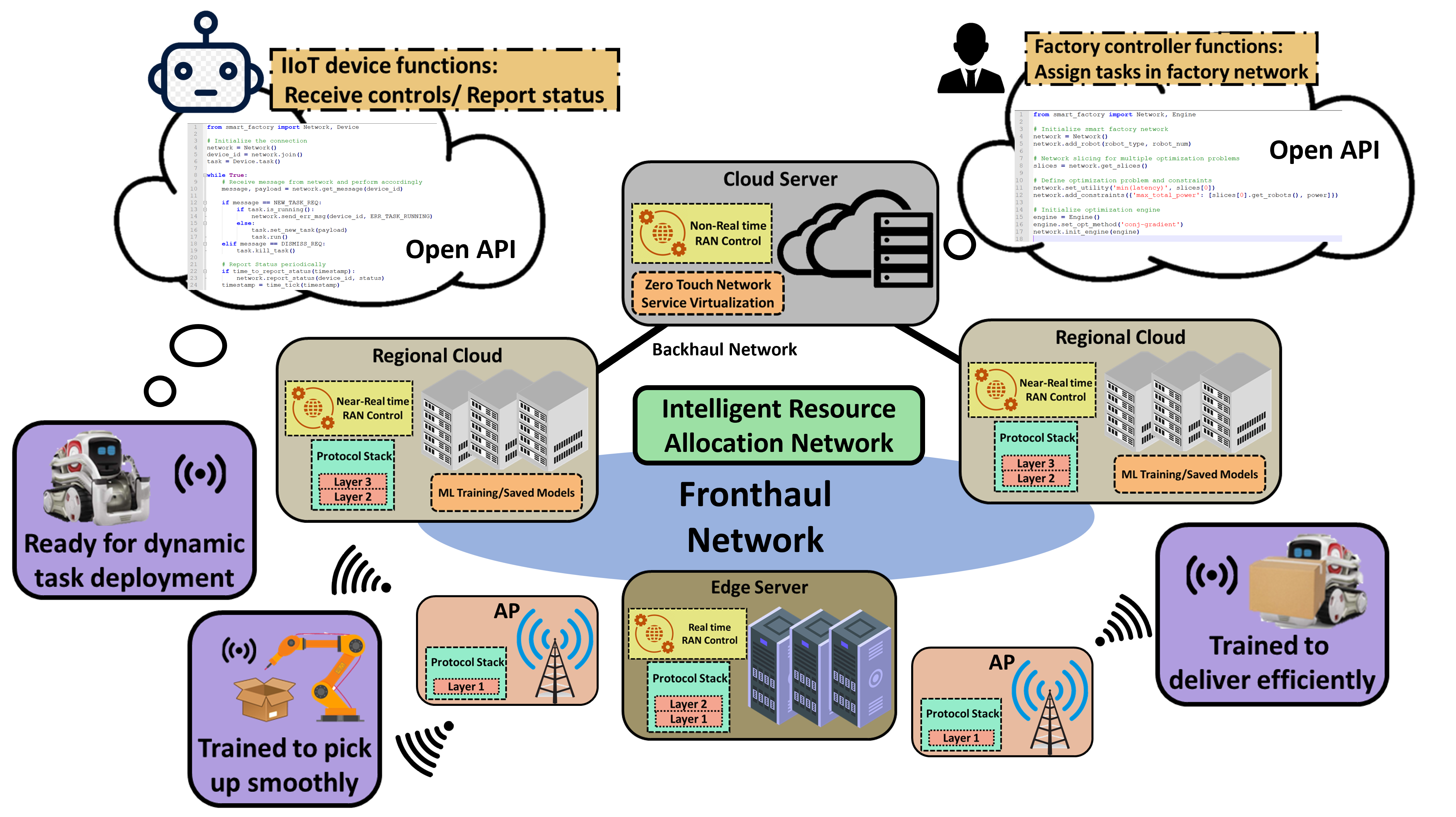}
    \caption{The illustration of the proposed zero-touch network for smart factory scenarios.}
    \label{fig:overview}
\end{figure*}

As shown in Fig. \ref{fig:sf_usecase}, to enable more complex operations to unbound the potential of smart factory, we envision large-scale and multi-task AI algorithms will lead the future of smart factory scenarios, achieving the final goal of Industry 4.0 to boost the productivity.
The considered scenarios is a complex field containing numerous human workers, robotic arms, and robot workers. The considered scenario also includes several computation and communication infrastructure to support the AI training and inferring among the aforementioned objects with different quality of service (QoS) requirements.
However, in a distributed manner, the training and inferring of advanced AI algorithms in complicate smart factory scenarios are not trivial.
First, as complex reinforcement learning (RL) and advanced deep learning algorithms \cite{Shiue.2018} are widely used to perform different tasks in smart factory scenarios, the distributed training of above algorithms is challenging \cite{DML}, often requiring significant communication and computation resources to support needed weight updating and exchanging among end-devices.
Second, after the aforementioned training stage, in the inferring stage, robots need to perform specific actions in time to collaborate with other machines or human workers. To do so, low-latency connectivity should be provided to ensure the inferring stage goes smoothly. 
As a result. from the communication perspective, there is an urgent need to develop a platform, offering strong backbone connectivity to support not only data-hungry training applications but also low-latency inferring applications. 
By doing so, the training and inferring of AI applications can be deployed in smart factory scenarios smoothly and effectively in a distributed manner. Note that this is especially important to smart factory scenarios as re-training may be frequently needed to adapt new tasks or new environments.
In order to offer the desired connectivity, the platform should be able to appropriately control underlying communication resources to satisfy different services, often requiring expert knowledge to do so. Even with expert knowledge to control the communication resources, to satisfy the computation demanding nature of AI applications, how to collaborate such platform to work with edge computing is still unsolved in literature \cite{EC, AIEC, AIEC2}.
In this paper, we propose zero-touch network to serve smart factory scenarios as illustrated in Fig. \ref{fig:overview}, allowing a central controller to fully exploit communication and computation resources to tackle distributed learning service requests with different requirements. Moreover, to realize next-generation networking system design, serverless framework \cite{Xie.2021} and open radio access network (O-RAN) structure are both considered in the platform implementations.
As a result, promising technologies, such as AL-enabled networking and edge computing \cite{AINetworking, AIEC, AIEC2}, can be deployed in the proposed platform effortlessly to enjoy network automation and reduced latency to further facilitate AI applications in both training and inferring stage.  

Specifically, with the developed platform, IIoT devices and infrastructure in smart factory field  could be virtualized, connected and managed together to aid the execution of AI applications.
On the one hand, an application programming interface (API) is designed for IIoT devices to report routes, current jobs, and battery status. Then the central controller can monitor and manage IIoT devices according to their current workloads, constraints and attributes (i.e., functionalities and locations), letting central controller to allocate training and inferring tasks to IIoT devices flexibly.
On the other hand, another API is offered to communication and computation infrastructure in field  to report the resources to central controller, letting central controller to designate specific resources to aid the above training and inferring tasks to achieve data rate or latency requirements based on the need of tasks.
To enable dynamic and efficient resource allocation capability, O-RAN architecture \cite{oran_website, Balasubramanian.2021} is introduced in our implementations, treating infrastructure in field  as different level units in O-RAN architecture. This not only enables efficient edge computing capability to aid AI applications, but also fits AI-enabled networking designs since the provided network function virtualization (NFV) and software-defined networking (SDN) allows us to execute and update advanced networking management schemes effortlessly. Furthermore, in the central controller, to process the reported information from IIoT devices and infrastructure efficiently, serverless framework is also considered in our implementations. With the dedicated function modules, the central controller can perform field -wide AI applications management and network and resource management for O-RAN units to facilitate AI applications in both training and inferring stages to achieve our final goal. 
Importantly, our implementations have three benefits to Industry 4.0-enabled smart factory scenarios. First, from the IIoT devices perspective, via our platform, we enable tasks allocations among robots to finish desired tasks to further reduce processing time, energy consuming, and other tasks. Moreover, new AI algorithms can also be trained in a distributed manner to tackle new tasks or changing environment. Second, from the networking system perspective, our platform can dynamically allocate communication and computation resources in the field to support above training or inferring tasks with various requirements. Finally, from the factory managers perspective, factory managers could simply utilize the given APIs to set up the parameters of the optimization scenarios, e.g., the machine learning (ML) algorithms, the optimization policies and constraints, and the number of each type of the end devices used in the scenarios, and the rest of the computing offloading, connection control, and service management would all be automated in our framework.
Based on the above services, factory managers could concentrate on the overall production performance and meet their changing demands immediately by maximizing the delivering throughput for target production lines and minimizing power consumption for others while keeping the dreadful details and knowledge of wireless connections, edge computing, artificial intelligence, coordination between heterogeneous computing resources, and radio resource management away.

The rest of this paper is organized as follows. Section \ref{sec_arch} presents the architecture of the proposed platform, and section \ref{sec_usecase} demonstrate current usages of our platform. Open research directions for zero-touch network in Smart Factory are specified in Section \ref{sec_opendir}, and finally the conclusion for this article is provided in Section \ref{sec_conclusion}.

\section{zero-touch network Platform Implementations}\label{sec_arch}

\subsection{Smart Factory Scenarios based on O-RAN Architecture}
\begin{figure*}
    \centering
    \includegraphics[width=\textwidth]{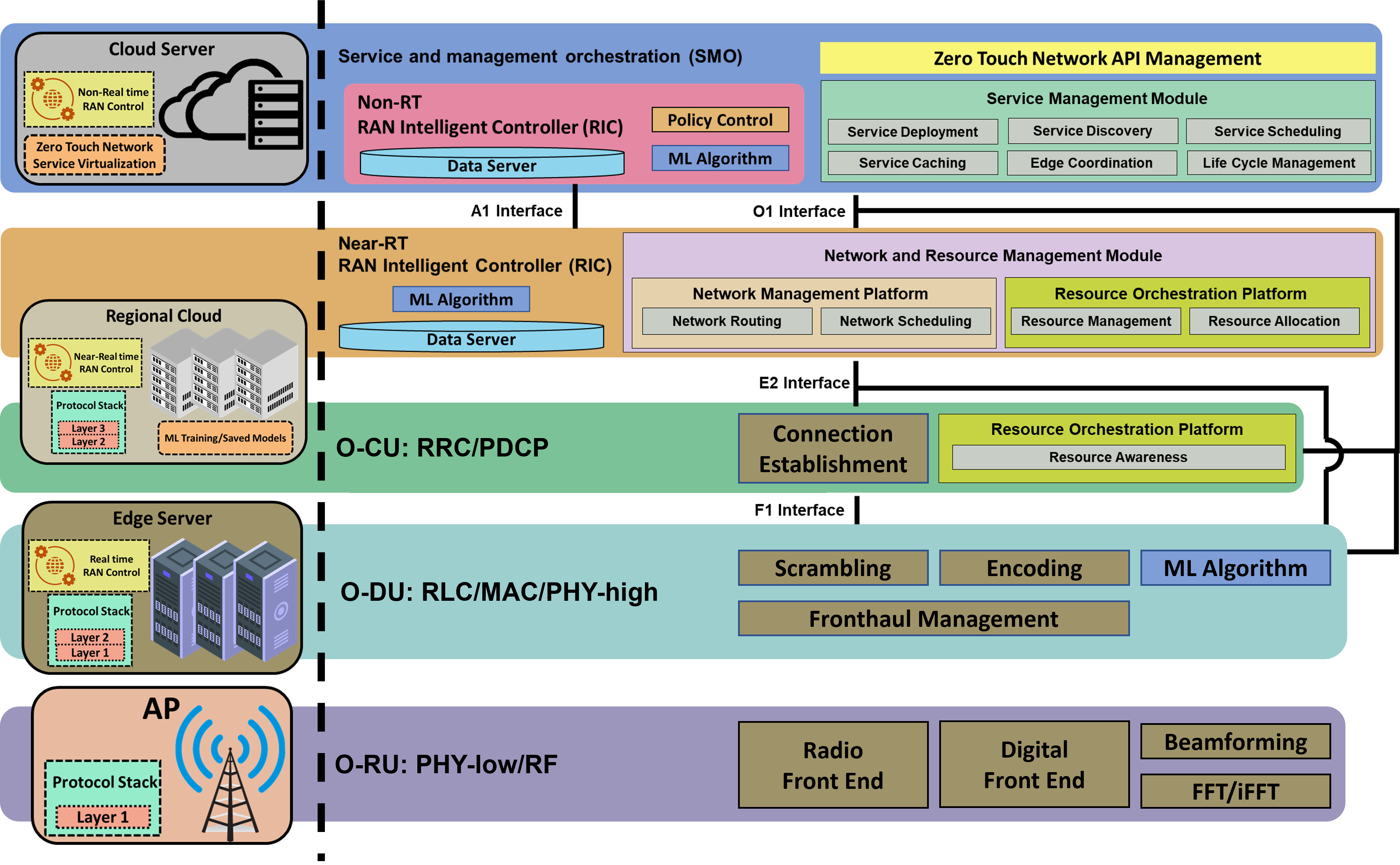}
    \caption{Architecture of our framework.}
    \label{fig:framework_arch}
\end{figure*}

In order to connect and manage all the IIoT devices and infrastructure in smart factory field effectively, hierarchical O-RAN structure is introduced in our platform implementations.
With O-RAN, we could utilize both the communication and computation resources in order to realize the connection of the IIoT devices and the virtualization of zero-touch network services.
On the one hand, in terms of communication infrastructure, there is the E2 device, composed of the open central unit (O-CU), open distributed unit (O-DU), and open radio unit (O-RU) in O-RAN, which represent the open interface-enabled CU, DU, and RU units in 5G protocol stack respectively, to provide different network layer functions. The O-RU hosts front end and functions in the lower part of the physical layer (PHY), such as beamforming and fast Fourier transform (FFT), the O-DU hosts the medium access control (MAC), radio link control (RLC) and the higher part of the physical layer functions, e.g., encoding and scrambling, and the O-CU hosts the radio resource control (RRC) and the packet data convergence protocol (PDCP) for connection management.
The O-CU could manage multiple O-DUs, while multiple O-RUs could be managed by one O-DU, and together those infrastructure assure the basic connections of all the devices in the O-RAN architecture.
Notably, the interfaces between them are open, namely that the O-CU, O-DU, O-RU components could be implemented by different manufacturers, letting the deployment of communication resources be flexible. Moreover, the O-CU, O-DU, and O-RU could be combined together in different ways for different scenarios, e.g., O-DU and O-RU could be integrated into one server, so as the O-CU and the O-DU, and even all of them could be integrated in the same server.
As shown in Fig. \ref{fig:framework_arch}, in the considered scenario, we deploy multiple access points (APs) as the O-RUs to provide the fundamental connections with IIoT devices, and design the edge server by employing the O-DU in it to manage the APs.
Specifically, APIs will be provided to robots and infrastructure to report status. Using the reported information, factory managers can assign learning tasks to appropriate robots and underlying infrastructure will be fully exploited to support the above tasks to facilitate large-scale AI applications in smart factory scenarios.
Furthermore, as O-RAN follows SDN and decouples the network control functions from the network architecture, the O-DU could utilize software-based network control loops executed within 1ms, and thus, trained machine learning models could be stored in the edge servers and network function applications on them could use these models to intelligently control the APs and the IIoT devices in ways such as radio scheduling and beamforming.

On the other hand, in terms of computation infrastructure, there are radio access network intelligent controller (RIC) and service and orchestration management (SMO) components in O-RAN, which are primarily adopted for software-based radio optimization and resources management for the E2 devices. Applications run on the RIC could be designed by the third party and utilized for configurable network functions. Moreover, the RIC could be separated to be near-real time RIC (nRT-RIC) and non-real time RIC (non-RT RIC) according to the network control loops execution time inside them.
To be more specific, the nRT-RIC controls load-balancing, interference detection and mitigation of tasks on the E2 devices, while the non-RT RIC operates within the SMO, handling the life-cycle management and configurations for all network elements, including the E2 devices and nRT-RIC, and it could provide policy-based guidance to the nRT-RIC to determine the RAN optimization actions.
To fully exploit these functionalities, we design the regional cloud server for combining the O-CU and the nRT-RIC and the cloud servers for the SMO and non-RT RIC as shown in Fig. \ref{fig:framework_arch}. These two types of servers are responsible for the ML training due to the fact that the training requires lots of computation power and space; therefore, they don't have to handle the connections between IIoT devices and are equipped with powerful servers.
Additionally, the cloud server supports the management of the APIs released to factory managers, including the service deployment, caching, scheduling, ...etc, and should be fulfilled by powerful computation facilities. 
As the SMO in O-RAN, the cloud server could orchestrate all the other components in the network, e.g., all the computation and communication resources, and fully virtualize all the network functions and devices operated in the Smart Factory.

\subsection{Serverless framework-enabled networking management}
To connect and manage all the IIoT devices, we utilize the network and resource management module in the serverless computing concept in our framework. 
In this platform, functionalities such as resource awareness, resource allocation/deallocation, network scheduling, and network routing control are introduced to efficiently orchestrate these IIoT devices.
To be more specific, this platform could keep track of the information from all the resources, including the battery status, current workloads, learnt models, and current positions, and deploy the suitable devices to train with new ML algorithms or execute certain tasks based on requirements from the service provided for the factory managers.
To realize these functions, robots and industrial arms will directly connect to the routers and edge servers, which would connect to the regional servers in a higher level, so that these IIoT devices can exchange information between each other and report their status in real time via given protocols in the framework.
And thus, functions in the network and resource management module are going to be implemented in these regional servers and edge servers, which represent the nRT-RIC, O-CU and O-DU, hosting for (near) real time control loops of the IIoT devices in our framework.
With the help of O-RAN architecture, interfaces between these components could be open and cross-platform, therefore guaranteeing the connections within the resource orchestration platform.

Last but not least, we introduce the service management module in the serverless computing concept for delivering the service to the factory managers.
Functions in this module include service deployment, service caching, and service discovery, and they will be implemented in the SMO, which is for high-level service and management orchestration of all other network functions.
Within the SMO, all resources and network status below would be virtualized, so that factory managers could determine the allocation of tasks among IIoT devices in an intuitive perspective.
As a result, zero-touch network could be realized with the integration of serverless computing and O-RAN architecture in our framework.
On one hand, with O-RAN, all the IIoT devices could be connected in an open way, which are beneficial for orchestrating between heterogeneous devices, and also be prone to design ML algorithms for AI training on specific tasks in the smart factory.
On the other hand, we could easily virtualize and manage all the IIoT devices by means of serverless computing due to the service management module and the network and resource management module.

\section{Learning-based Networking System Enhancement}\label{sec_usecase}

\begin{figure*}
    \centering
    \includegraphics[width=\textwidth]{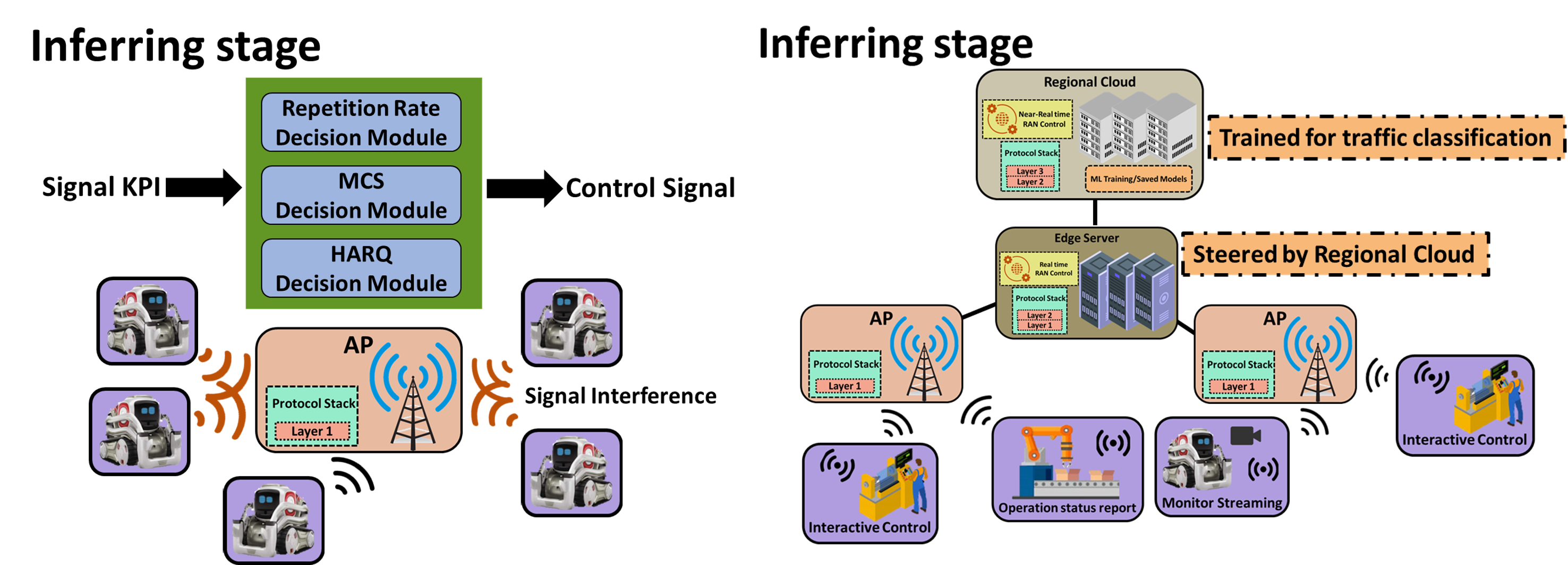}
    \caption{Inferring stages of our two RL-based enhancements. Upper: Transmitter Reconfiguration; Lower: Traffic Steering.}
    \label{fig:use_case_arch}
\end{figure*}

Due to the flexibility of the developed platform, we can further introduce powerful algorithms into zero-touch network to realize AI-enabled networking system to enjoy the provided benefits. Specifically, we present two learning-based enhancements shown in the Fig. \ref{fig:use_case_arch}, which can be used to further improve QoS in the proposed platform in this section. First,   
we introduce our RL-based transmitter reconfiguration algorithm to perform radio resource management for smart factory scenarios. Second, we explain our ML-based network traffic steering algorithm to provide QoS guarantees to specific type of data traffic.
Both algorithms can be used in our platform to facilitate AI applications in training and inferring stages.

\subsection{RL-based Transmitter Reconfiguration}\label{sec_aic}
\subsubsection{Problem Statement}
In the considered smart factory scenario, in a downlink procedure, we assume that $N_{UE}$ robots are served by an AP concurrently. Due to the mobility of robots, the wireless conditions of each link will be highly time-variant. Consequently, the AP should be able to adjust transmission parameters, adapting new environment to realize dynamic and efficient resource allocations automatically \cite{MCSbyRL}.
Hence, our goal is to design an algorithm to perform transmitter configuration adjustments based on reported key performance indicators (KPIs) from robots automatically. As each user is required to report KPIs to AP periodically, our algorithm can be used to accustom time-variant wireless conditions and reconfigure transmitters accordingly.
Specifically, to match the functionality of current commercial used communication system, we define the reported KPIs of each link as signal-to-noise ratio (SNR), block error rate (BLER), latency, throughput as the input of developed algorithm. As for the output of the developed algorithm, decision making in terms of modulation and coding scheme (MCS), repetition rate, and maximum hybrid automatic repeat request (HARQ) designs is expected to be executed to realize transmitter reconfiguration according to faced environment.


\subsubsection{Proposed Algorithm}
Originally, to solve the desired decision making problem, searching algorithms among a fixed-size finite set can act as solutions.
However, due to the limited responding time nature of smart factory scenarios, computationally demanding algorithms are prohibited for the whole resource allocation procedure. Consequently, pre-designed look-up tables are widely used in current commercial communication systems but compromise the achieved performance. 
Alternatively, we propose our reinforcement learning (RL)-based transmitter reconfiguration solution. Specifically, in the AP side, we deploy $N_{UE}$ agents to let each agent be responsible for a downlink connection. Then reported KPIs are set as the state to each agent and each agent can adjust MCS, repetition rate, and HARQ as actions independently to minimize achieved BLER.
Our designs enables two important features to smart factory scenarios.
First, we realize adaptive transmitter reconfiguration in a data-driven manner even without labels. 
To explain, 
in the considered scenario, taking the employed commercial system-level simulator as an example, in each downlink connection, up to $576$ options can be adjusted to perform joint decisions of MCS, repetition rate, and HARQ in transmitter. As $N_{UE}$ robots are severed by the same AP simultaneously, the number of options will expand to $(576)^{N_{UE}}$, making it almost impossible to generate labels via existing searching algorithms. 
Alternatively, via the proposed RL-based algorithm, our agent can perform efficient searching by the provided exploring ability to improve pre-defined reward function, finish training even without the existence of pre-defined labels. Note that this is important in smart factory scenarios as frequently re-training may be required to adapt new tasks or new environment in practice.
Second, we realize more complex and flexible transmitter reconfiguration based on multi high-level KPIs.
To explain, traditional MCS design only takes SNR as input for decision making. Alternatively, through our design, the proposed algorithm can extract more information from multi KPIs to aid the decision making. Moreover, the reward function can be set as different performance metrics to improve desired QoS indicators, letting us design customized communication system to satisfy different needs of enhanced Mobile Broadband (eMBB), Ultra Reliable Low Latency Communications (URLLC), and massive Machine Type Communications (mMTC) in smart factory scenarios. Furthermore, the proposed multi-agent solution can be extended easily to perform more complex transmitter reconfiguration including power allocation and beamforming design to joint link designs in the transmitter side. The training and inferring stages are also illustrated in the upper part of Fig. \ref{fig:use_case_arch}.

\subsubsection{Testing Results}
\begin{figure}[b]
    \centering
    \includegraphics[width=3.4 in]{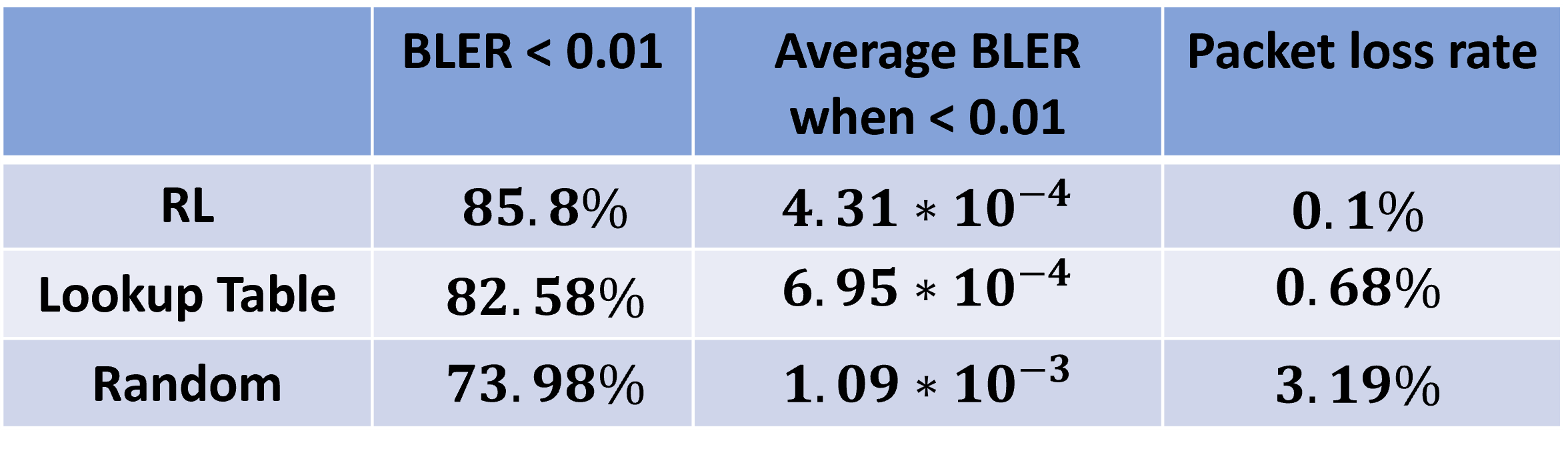}
    \caption{Testing results of the ML-based transmitter reconfiguration.}
    \label{fig:tran_reconf_result}
\end{figure}

We develop a digital-twin site that allows us to perform model training and support performance evaluation for smart factory. Specifically, following O-RAN structure, during training phase, the designed agents will connect to a system-level communication system through E2 interfaces to interact with environment. In the testing phase, the trained agent will be placed in the nRT-RIC to perform dynamic and efficient transmitter reconfiguration. In the evaluation results reported in this paper, we set $N_{UE}$ as five to let an AP serve five robots simultaneously. All robots are moving with the speed of three km/hr to random direction to present the mobility in smart factory scenarios.
We compare our results with two baseline solutions: random selection-based and lookup tables-based transmitter reconfiguration algorithms to prove the provided superiority. Note that due to the large dimension of the action space and the online nature of the digital-twin site, exhaustive searching or other searching algorithms cannot be employed to solve the interested problem.  
In Fig. \ref{fig:tran_reconf_result}, one can notice that the proposed algorithm outperforms random selection-based and common used lookup table-based transmitter reconfiguration in different performance metrics. This is because our algorithm can reference more KPIs to perform decision making and consequently enjoy the superiority. Note that lookup table-based algorithm is the standard solution in conventional communication systems and our algorithm can replace it in smart factory scenarios to perform near real-time transmitter reconfiguration designs. Our algorithm can efficiently supervise communications among multiple robots without any participation of factory managers.

\subsection{Machine Learning-based Traffic Steering}\label{sec_fisc}

\subsubsection{Problem Statement}
In the considered smart factory scenario, the data traffic is actually bustling. That is, several types of services are operating in networking systems at the same time as shown in the lower part of Fig. \ref{fig:use_case_arch}, such as video streaming services for camera monitoring usage, interactive data services for controlling signal usage and bulk data transfer for AI data and weights exchanging usage.
Moreover, different type of services have different QoS requirements to be satisfied \cite{Wang.2016}. To fulfill the QoS requirements of latency-sensitive applications in smart factory scenarios with limited networking resources, we further develop traffic steering algorithm presented in this section. Specifically, our goal is to design an algorithm to perform intelligently automatic traffic steering based on KPIs in network traffic flow level. Since it only requires network KPIs as input, our algorithm can be performed in routers without modifying any existing protocols. 
In detail, we extract KPIs from network flows including timestamp, downlink buffer bytes, and employed network protocol as the input of developed algorithm to perform service classifications as the first step. Then based on the classified results, limited physical resource blocks (PRBs) will be assigned to latency-sensitive services to satisfy the QoS requirements of those applications \cite{TSbyRL}. 

\subsubsection{RL-based traffic steering}
In the proposed algorithm, network traffic flows will be classified into four different types of services, voice/video conference, interactive data, streaming service, and bulk data transfer, since the network requirements of different services vary significantly.
The packets in the same services type will be aggregated into the same queue, and sorted in order according to the arrived time to routers.
After the classification, the algorithm will check if the current status of network communication resources are enough to fulfill all the QoS requirements of received packets in the queues. If the PRBs are scarce compared to the demands, the second step will be executed to perform packet scheduling to guarantee the QoS of latency-sensitive applications.
Specifically, PRBs will be reserved for the packets from the target services, guaranteeing their performance first. After satisfying those target services, the packets from different queues will start to be served by assigning the rest of the PRBs to those applications.
Note that our algorithm can be extended and customized to support more services type and improve desired QoS indicators more than latency.


\begin{figure*}
    \centering
    \includegraphics[width=\textwidth]{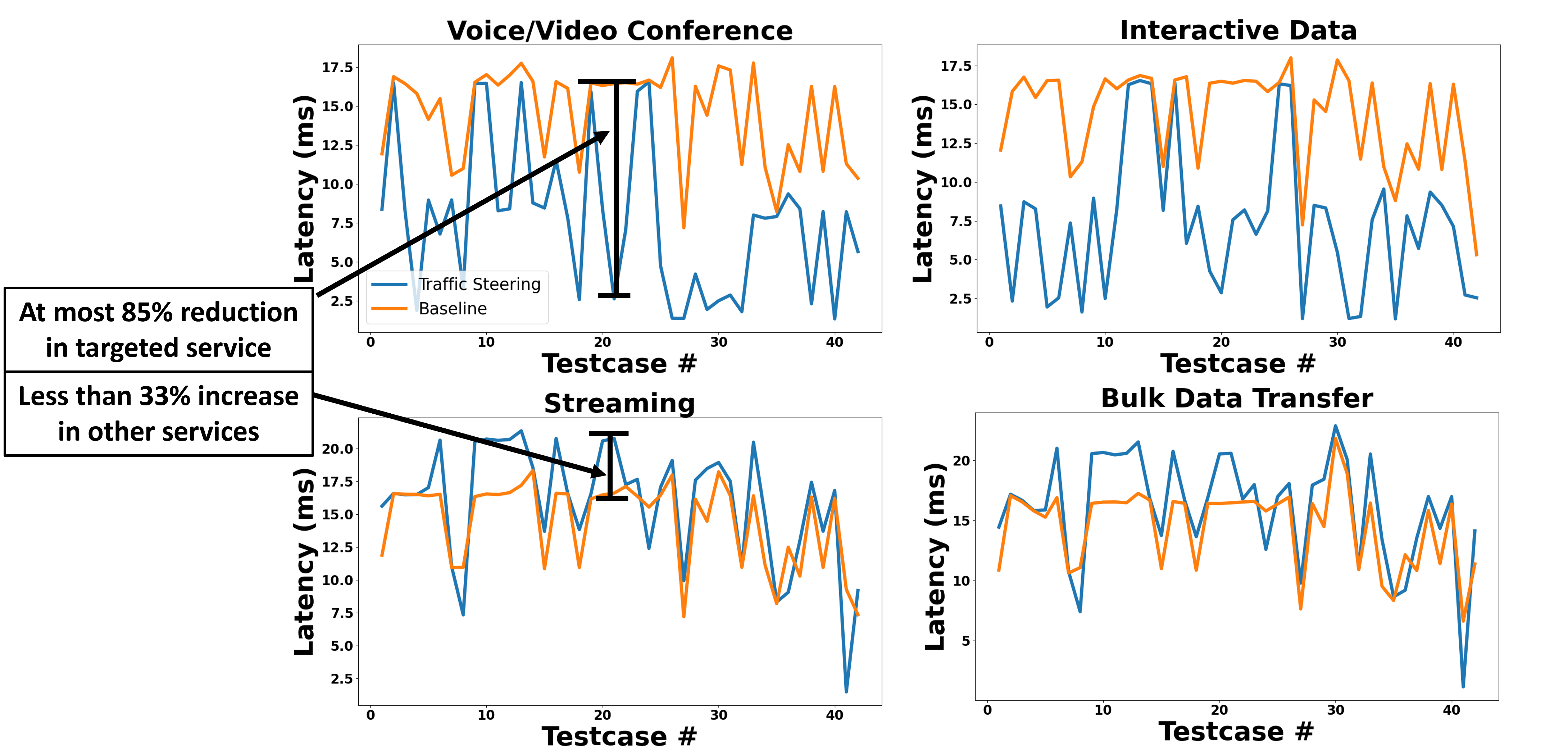}
    \caption{Testing results of the ML-based traffic steering.}
    \label{fig:traf_steer_result}
\end{figure*}

\subsubsection{Testing results}
We utilize Colosseum dataset \cite{Bonati.2021} to train and test our ML-based traffic steering algorithm. In the Colosseum dataset, there are four APs operating in dense urban scenario and each AP serves ten UEs simultaneously to obtain the network traffic data.
To test our algorithm in a insufficient network resources environment, we set up only three PRBs are available for transmission, which is not enough for all types of services. 
We focus on minimizing the average latency for the voice/video conference and interactive data services, and we reserve two PRBs for those services.
That is, whenever there are packets for these two services that need to be served, the reserved PRBs will only be utilized for that purpose.
The rest PRB will be used to serve for all other packets in order according to the arrived time. With this method, services other than the targeted service can still be sent out smoothly.
We compare our results before and after adopting the traffic steering algorithm as shown in Fig. \ref{fig:traf_steer_result}.
One can notice from the results that the average latency of voice/video conference and interactive data service is improved significantly to promise the latency quality, while other two services only get a little higher latency than usual because of the shortage of PRBs. Since these two service do not demand low latency, our algorithm can aid both bustling data traffic and limited networking resources scenarios to satisfy different QoS requirements at the same time.

\section{Future Research Directions}\label{sec_opendir}
To meet the urgent need of smart factory scenarios, the development of next-generation zero-touch network needs more attentions of experts from both academia and industry.We list three crucial research directions to integrate existing technologies into the provided platform to enable advanced Industry 4.0 in smart factory.

\subsection{Integration with the Legacy Tools and Codes}
The amount of existing solutions and codes for conventional Industry 3.0 represents a vast investment of time and money for developers and manufacturers to manage the production lines and integrate the corresponding hardware and software in an efficient and operable way. Accompanied by them, there are countless legacy tools for debugging and monitoring all the devices in the factory.
As serverless computing and O-RAN architecture possess different granularity than conventional ways, the most important question is how to leverage the legacy solutions and decouple them to fit into the new architecture so that the new economics could be fully exploited.
Since the devices would all be connected together to form an extreme wider network and the codes would be separated into small pieces resided in different components, how to extend the existing tools to debug and neatly control the system could be challenging. And new approaches are necessary for virtually assemble the serverless pieces into an easier and understandable way for system management.

\subsection{Balance between Training Speed and Accuracy}
Owing to the usage of multiple AI/ML models for non/near real-time RAN optimization in our framework, a balance should be found between the training time and accuracy of the AI/ML models.
Transfer learning aims to leverage the knowledge of one task to solve other related tasks. To be more specific, data collected to train a ML model, e.g., a series of signals transferred between the edge server and the IIoT devices, are used to train the AI to realize automatic signal interference control, which could also be used to support the training for another ML model, e.g., the intrusion detection for the malicious attacks models.
With transfer learning, fast training process of multiple ML models could be reached; nonetheless, the major problem of it would be to identify how to utilize the transfer learning so that negative effects on the ML model performance could be avoided. Hence, more research topics are required in this direction to utilize the capabilities of the transfer learning in our framework.


\subsection{Cybersecurity}

Last but not least, cybersecurity of the zero-touch network in Smart Factory could never be forgotten. Since in zero-touch network, all resources are connected together, information leakage may occur if too much details of the data are transmitted on the air, which are prone to evesdropping. As a result, the design of the algorithms that provide well enough performance while keeping the requirements of the amounts of data as limited as possible demand to be proposed to avoid this kind of problems.
Another part is the design of the user APIs, that is, how to guarantee the services to not be cracked to open backdoors for malicious users. Consider the complexity of the whole system, holes could be inevitable if not being evaluated thoroughly. And thus the verification models and processes of the zero-touch network are vital and need more advanced research.

\section{Conclusion}\label{sec_conclusion}
This paper provides a comprehensive network system for smart factory to execute AI applications in training and inferring stages efficiently. The proposed approach is especially suitable for intelligent manufacturing to enable large-scale AI applications deployment in practice. With the provided flexibility, we demonstrate that our system is also compatible with popular AI-enabled networking algorithms to enjoy the provided benefits in terms of improved QoS performance. Lastly, we discuss future research directions to next-generation zero-touch networks for smart factory scenarios, encouraging more researchers to contribute their effort in this direction.

\bibliographystyle{IEEEtran}
\bibliography{ref}

\end{document}